\documentclass[runningheads]{llncs}
\usepackage[T1]{fontenc}
\usepackage{graphicx}
\usepackage{xcolor}
\usepackage{multirow}
\usepackage{url}

\begin{document}
\title{Inverse Scene Text Removal}
%
%

\author{
Takumi Yoshimatsu
\orcidID{0009-0002-9259-8275} 
\and Shumpei Takezaki\orcidID{0009-0005-3963-315X} \and Seiichi Uchida\orcidID{0000-0001-8592-7566}
}
\authorrunning{T. Yoshimatsu et al.}
%
\institute{Kyushu University, Fukuoka, Japan\\
\email{\{takumi.yoshimatsu, shumpei.takezaki\}@human.ait.kyushu-u.ac.jp 
uchida@ait.kyushu-u.ac.jp
}}
\maketitle              
\begin{abstract}
Scene text removal (STR) aims to erase textual elements from images. It was originally intended for removing privacy-sensitive or undesired texts from natural scene images, but is now also applied to typographic images. STR typically detects text regions and then inpaints them. Although STR has advanced through neural networks and synthetic data, misuse risks have increased. This paper investigates Inverse STR (ISTR), which analyzes STR-processed images and focuses on binary classification (detecting whether an image has undergone STR) and localizing removed text regions. We demonstrate in experiments that these tasks are achievable with high accuracies, enabling detection of potential misuse and improving STR. We also attempt to recover the removed text content by training a text recognizer to understand its difficulty.
Our datasets and model are available at: \url{https://github.com/takumi-yoshimatsu/ISTR}

\keywords{Scene text removal\and Target region detection \and Tamper detection.}
\end{abstract}
\begin{figure}[t]
    \centering
    \includegraphics[width=0.70\columnwidth]{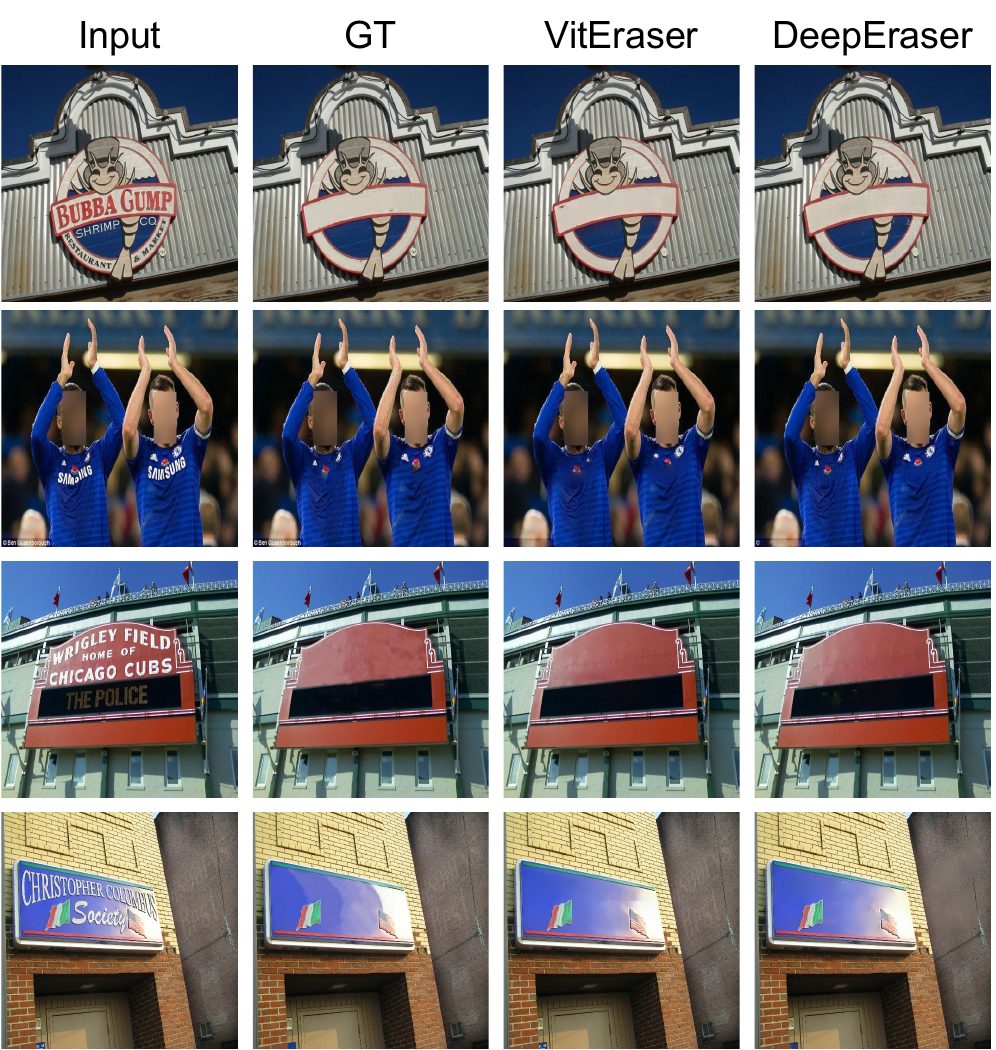}
    \caption{Results by state-of-the-art (SOTA) STR techniques, ViTEraser~\cite{peng2024viteraser} and DeepEraser~\cite{feng2024deeperaser}.}
    \label{fig:str_example}
\end{figure}

\begin{figure}[t]
    \centering
    \includegraphics[width=\columnwidth]{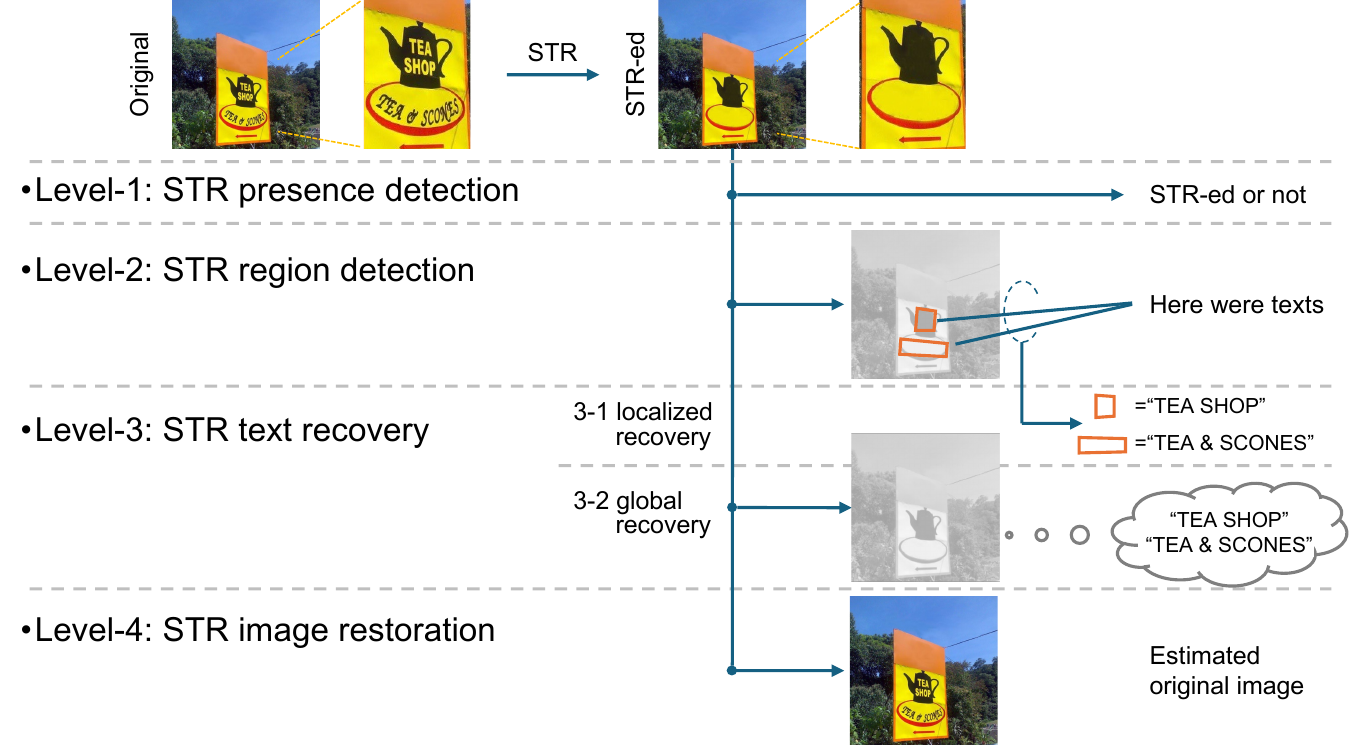}
    \caption{Four levels of ISTR.}
    \label{fig:ISTR-steps}
\end{figure}

\section{Introduction\label{sec:intro}}
Scene text removal (STR) aims to inpaint text regions in images. STR was originally intended to remove privacy-sensitive texts (e.g., car license plate numbers or house numbers) or otherwise undesired texts from natural scene images prior to their publication. Recently, STR has also been applied to typographic images, such as posters or digital advertisements, to facilitate subsequent retouching (e.g., \cite{shimoda2023iccv}). 
\par

Generally, STR involves two core functions: detecting text regions and inpainting (i.e., filling in) those detected regions. These two functions can be performed in a step-by-step process or in a single end-to-end framework. As reviewed in the next section, numerous STR methods have been proposed, showing continual performance improvements through advances in neural network architectures. Fig.~\ref{fig:str_example} shows STR results by two state-of-the-art (SOTA) STR techniques, ViTEraser~\cite{peng2024viteraser} and DeepEraser~\cite{feng2024deeperaser}. For human eyes, it is almost impossible to detect any traces of STR in these results.\par
Unfortunately, these advanced STR techniques pose a growing risk of misuse, such as concealing crucial information. For example, STR could remove copyright notices or other textual evidence in images or serve as a preliminary step before more malicious image tampering. Historically, negative messages on placards unfavorable to a government have often been physically removed; now, STR can make such manipulation considerably simpler and more accurate.
\par

This paper investigates \emph{Inverse} STR (ISTR), which aims to find out some traces of STR from ``STR-ed'' images. More specifically, as illustrated in Fig.~\ref{fig:ISTR-steps}, ISTR can be defined as four tasks with different levels according to their expected difficulty:
\begin{itemize}
    \item {\bf STR presence detection} (level-1 ISTR) aims to simply classify whether STR was applied to an input image. 
    \item {\bf STR region detection} (level-2 ISTR) aims to localize the removed text regions. 
    \item {\bf STR text recovery} (level-3 ISTR) aims to estimate the removed text contents. 
    \item {\bf STR image restoration} (level-4 ISTR) aims to restore the original images before STR.
\end{itemize}
Precisely speaking, Level-3 ISTR is further divided into two sub-levels, Level-3-1 and Level-3-2, as shown in Fig.~\ref{fig:ISTR-steps}. Level-3-1 assumes that the removed text regions are already localized as the solution of Level-2 and aims to ``recognize'' the removed texts in the individual regions. Level-3-2 is the case where no text regions are available before recovery and aims to ``seek and recognize'' the removed texts in the whole image. 
\par

This paper aims to experimentally verify to what extent the above four ISTR levels are achievable against the SOTA STR techniques. To summarize our experimental conclusion in advance, we can detect Level-1 and Level-2 with high accuracy even against the STR-ed images by SOTA STR models. On the other hand, Level-3 (particularly the simpler Level-3-1) appears nearly impossible, even if we attempt text recognition of removed regions using the latest technologies. Consequently, we consider the higher-level Level-4 to be intractable at present.\footnote{We regard Level-4 as intractable because, without recognizing the removed text, it is difficult to fully restore it at a readable level. However, ignoring text readability, partial recovery of some letters may be possible through an approach that does not explicitly involve text recognition. We leave this issue as future work.}
\par

It is important to note that there is no need to be pessimistic if we are unable to verify that all these levels are achievable. If we verify that the first two levels are achievable, we understand that ISTR is already helpful in detecting potential misuse of STRs. Moreover, the achievements in Level-1 and 2 also indicate that ISTR may be useful to further improve STR by acting as a discriminator in a generative adversarial network (GAN), pushing the generator (i.e., STR) to produce ever more convincing outputs (i.e., more perfect text removal).
\par

\section{Related Work}
\subsection{Image Inpainting}

Inpainting is a technique for erasing part of an image while inconspicuously filling the removed region. It is already introduced in commercial software such as Adobe Photoshop, and widely used to protect subjects' privacy. Inpainting methods can be broadly categorized into non-learning-based approaches and learning-based approaches. Non-learning-based methods include classical approaches (e.g., \cite{1572261550764376192}) that smoothly fill in specified regions by blending information, such as the weighted average of nearby pixels. Another well-known approach is patch matching \cite{barnes2009patchmatch}, which searches for similar patches surrounding the target area and embeds them to create natural-looking completions. However, these non-learning-based approaches typically struggle with larger or more complex regions.
\par
By contrast, learning-based methods use neural networks to transform images. Although these approaches require training data, they can yield more powerful inpainting results. For example, Context Encoder \cite{pathakCVPR16context} introduced an encoder-decoder architecture with an adversarial loss for learning inpainting. A more recent example is inpainting with Latent Diffusion \cite{rombach2022high}, which compresses images into a low-dimensional latent space, adds noise, and then gradually removes that noise to reconstruct the original image. This approach can fill removed regions in harmony with the surroundings, even for complex structures. LaMa \cite{suvorov2022resolution} further leverages fast Fourier convolution to incorporate broader contextual information, enabling natural and consistent inpainting over larger areas compared to traditional CNN-based methods.
\par

We focus on STR, rather than traditional inpainting for general object removal. 
Because the former is an image transformation technique using end-to-end machine learning, whereas the latter is an image completion technique to fill a (often manually) removed region.

\subsection{Text Region Detection}
Text region detection, or scene text detection, is a type of object detection task that aims to locate text areas within images (particularly scene images). Although it used to be considered a challenging problem, recent advances in machine learning and synthetic data generation techniques have led to dramatic progress, allowing for highly accurate detections today. Methods for text region detection can be broadly classified into segmentation-based and regression-based approaches. In segmentation-based methods, the algorithm internally predicts whether each pixel or small segment belongs to text, and then aggregates these predictions to detect the text regions~\cite{baek2019character,liao2020real,long2018textsnake}. By contrast, regression-based methods treat text as a single object and directly predict the bounding box or polygon coordinates of the text~\cite{hou2020ham,liao2017textboxes,zhou2017east}. Notably, text regions are not always well-represented by simple rectangular bounding boxes. To address this, some methods, such as TextBPN-Plus-Plus~\cite{zhang2023arbitrary}, have been developed to detect arbitrarily shaped text regions. 
\par


%

\subsection{Scene Text Removal (STR)}
STR models can be divided into two types. The first type is two-step STR, where texts in the input image are detected and then the detected region is inpainted to remove the texts. Qin et al.~\cite{qin2018automatic} introduced a GAN-based framework with one encoder and two decoders, aiming at both text localization and background completion. MTRNet~\cite{Tursun_2019} improves performance by using an auxiliary mask that coarsely indicates text locations, aiding the Inpainting process. DeepEraser~\cite{feng2024deeperaser} adopts a recursive architecture that repeatedly erases text regions specified by a mask image. Note that this type can select the target texts to be removed by feeding the target regions to the second inpainting step~\cite{feng2024deeperaser,Mitani_2023_BMVC}.
\par
The second type is end-to-end STR, where text regions are directly inpainted without any explicit text detection process. Scene Text Eraser~\cite{nakamura2017scene}, which is the first machine-learning-based STR, employs a U-Net to convert the original image into a text-less image. EnsNet~\cite{zhang2019ensnet} uses a CGAN~\cite{cgan} framework in combination with four different loss functions to further boost text removal performance. SAEN~\cite{du2023modeling} deviates from bounding-box-based methods by predicting text strokes inside the network and removing text accordingly. ViTEraser~\cite{peng2024viteraser}, which is one of the SOTA STR models, leverages Vision Transformer modules in both the encoder and decoder to capture long-range dependencies, enabling more natural and consistent background restoration.
\par

We focus on complete STR in this study. We believe that partial STR is similarly solvable, as our detectors fundamentally rely on removal artifacts, which consistently appear with partial removal as well. This is also supported by the Grad-CAM~\cite{selvaraju2017grad} results in Section 3.3.

\section{STR Presence Detection\ (ISTR Level-1)} 
\subsection{STR Presence Detection as a Binary Classification Task}
The goal of this section is to determine whether an input image has undergone STR (i.e., whether text was removed). This corresponds to ISTR Level-1 in Fig.~\ref{fig:ISTR-steps}; in the example shown in this figure, where the texts (``TEA SHOP'' and ``TEA \& SCONES'') are removed by STR, the model should classify the image as ``STR-ed.'' In other words, it is a binary classification task that distinguishes images with STR from those without.
\par

If we can prove that STR presence detection achieves high accuracy even against SOTA STR methods, we can use this ISTR for various tasks. For example, we can identify whether text content in an image has been maliciously modified or retouched by STR. By knowing the presence of these changes, we can take a critical step toward verifying the integrity of the document.
\par

We adopt the standard binary classification framework for STR presence detection. Specifically, we prepare a dataset consisting of STR-ed images (as positive instances) and images that originally had no text (as negative instances), then train a model to distinguish these two classes. 
Note that this study assumes that the STR process aims to remove all text from the image, as most STR approaches attempt to eliminate all visible text. 
For this reason, we restrict the negative instances to images without any text. If negative instances contained text, the model could simply detect the presence of text as a cue for the negative class, thus trivializing the classification.

\subsection{Experimental Setup for ISTR Level-1}

\begin{figure}[t]
    \centering
    \includegraphics[width=0.6\columnwidth]{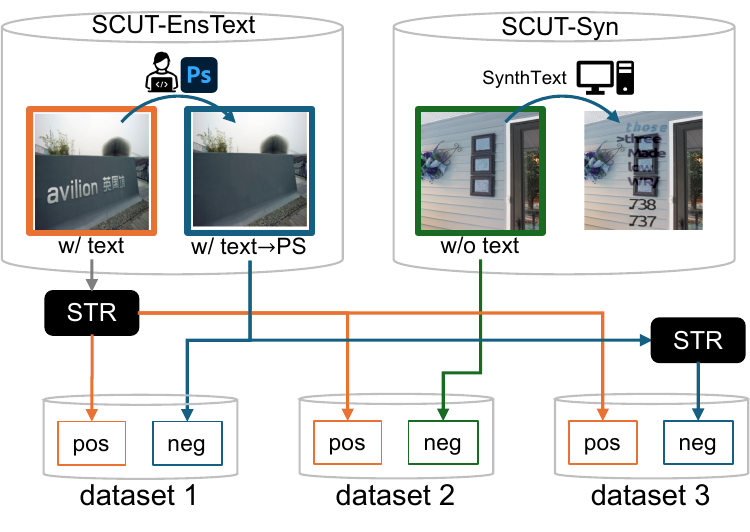}\\[-4mm]
    \caption{Three datasets for ISTR Level-1.}
    \label{fig:level1-datasets}
    \vspace{-4mm}
\end{figure}

For ISTR Level-1, we prepared three datasets (Datasets~1, 2, and 3) to more thoroughly evaluate the possibility of STR presence detection. Fig.~\ref{fig:level1-datasets} illustrates how these datasets were prepared. Dataset~1 is a baseline dataset created from a public STR dataset, SCUT-EnsText~\cite{Erase}; however, as noted below, we suspect there may be invisible biases in the evaluation with Dataset~1. Therefore, we prepared the other two datasets to ensure the robustness of our conclusions. Using the three datasets independently, we trained three different classifiers and evaluated them. Note that all images in the datasets are $512\times 512$ pixels in size.

\subsubsection{Dataset 1}
We use the SCUT-EnsText dataset~\cite{Erase}, which comprises 2,749 images containing text and the same number of images where the text was removed using Adobe Photoshop. We apply a STR model to the former images to obtain positive (i.e., STR-ed) instances. Meanwhile, the latter images serve as negative instances (i.e., images not processed by STR). Consequently, we obtain 2,749 positive and 2,749 negative instances for training our binary classification model. SCUT-EnsText also includes 813 test images, from which we similarly generate 813 positive and 813 negative test instances.



\subsubsection{Dataset 2}
One concern is that the negative instances in Dataset 1 are still ``artificial'' text-free images created with Photoshop. If there is any invisible artifact left by the Photoshop operation, the classifier might simply learn to distinguish between ``Photoshopped vs. not Photoshopped’’ rather than recognizing STR-ed or not. To address this concern, we prepared another dataset whose negative instances are {\em purely} negative images that never contained text. Specifically, we randomly select 2,749 images from 8,000 images of the text-free training set of SCUT-Syn~\cite{Erase} for the negative training instances. Moreover, we use 800 images from text-free test set of the dataset as the negative test instances.
\par

\subsubsection{Dataset 3}
Another concern is that the negative instances of Dataset~1 have not passed any STR model, i.e., neural network layers, whereas the positive instances have passed. If the layers in the STR model leave any invisible ``fingerprint'' within the STR-ed image, the classifier might focus on the fingerprint. To address this concern, we also prepared another dataset whose negative instances were prepared by applying STR to the original 2,749 negative images in SCUT-EnsText. Since the original negative images do not contain any text (by the operation with Photoshop), their STR-ed version is the same as the original for human eyes.\footnote{As discussed later, STR, especially ViTEraser, sometimes ``over-removes'' non-text regions. This makes a negative image like a positive image, and thus, such an over-removed negative image tends to be misrecognized as a positive image.}

\subsubsection{STR models}

We employ two SOTA STR models: DeepEraser~\cite{feng2024deeperaser} and ViTEraser~\cite{peng2024viteraser}. The former is a SOTA of two-step STR models, whereas the latter is a SOTA of end-to-end STR models. DeepEraser needs mask images to identify non-text regions so that its inpainting step does not modify the masked region. The mask images are generated from the text region provided in SCUT-EnsText as ground-truth. Both of these STR models are trained on the SCUT-EnsText dataset described in the previous section. Among several versions of ViTEraser (i.e., Tiny, Small, Base), we employed ViTEraser-Base because it achieved the best erasing performance among them. 

\subsubsection{Classification model}

As the classification model, we employed ResNet50, pretrained on the ImageNet dataset. We trained the model using the positive (``STR-ed'') and negative (``not STR-ed'') instances from Datasets~1, 2, or 3. Specifically, for each dataset, we randomly split 2,199 images (i.e., 4,398 including both positive and negative) for training and 550 images (i.e., 1,100 total) for validation from the dataset's training instances. We then used the classifier weights from the epoch that achieved the highest accuracy on the validation set as the final model. The learning rate was set to 0.0001, the batch size to 64, and the number of epochs to 50; Adam was used as the optimizer.

\begin{table}[t]
 \centering
 \caption{STR presence detection accuracy (\%) $\uparrow$ on the test images . ResNet50 was used as the detection model. See Fig.~\ref{fig:level1-datasets} for the details of the three datasets. text: images containing text. w/o text: images that never contained text. PS: Photoshop for manual text erasing. \label{table:STR-train-acc}}
  \begin{tabular}{rr|c|c|c}
   \hline
      \hphantom{hogehoge}& & Dataset 1 & Dataset 2 & Dataset 3 \\ \cline{2-5}
   \hphantom{hogehoge}&Pos. & w/ text $\to$ STR & w/ text $\to$ STR & w/ text$\to$ STR \\ \cline{2-5}
   &Neg. & w/ text $\to$ PS  &  w/o text & w/ text$\to$PS$\to$STR \\ \hline \hline
    \multicolumn{2}{l|}{ViTEraser~\cite{peng2024viteraser}}  & 98.89 & 99.93 & 93.05 \\
    \hline
    \multicolumn{2}{l|}{DeepEraser~\cite{feng2024deeperaser}}  & 97.97 & 95.91 & 97.29 \\
   \hline
  \end{tabular}
\end{table}

\subsection{Experimental Results of ISTR Level-1}
\subsubsection{Quantitative evaluation}
Table~\ref{table:STR-train-acc} shows the accuracy of STR presence detection on Dataset~1 when using two state-of-the-art (SOTA) STR models. Surprisingly, the classifier achieves near-perfect accuracy of around 98\%. This strong result demonstrates that ISTR Level-1 is highly achievable, even against advanced STR approaches. In other words, these SOTA STR models leave subtle ``invisible'' traces in their output images. Later, in our Level-2 experiments, we will confirm this phenomenon again, showing that removed text regions also carry such traces.
\par

The second column of Table~\ref{table:STR-train-acc} presents the detection accuracy on Dataset~2. Both STR models exceed 95\% accuracy, indicating that the classifier is not merely learning to distinguish ``Photoshopped vs. non-Photoshopped’’ images. Instead, it reliably identifies STR-processed images, suggesting that these images contain a unique ``fingerprint’’ left by the STR operation. This finding reinforces the idea that text removal --- even at a high level of quality --- still imparts detectable artifacts.
\par

The third column of Table~\ref{table:STR-train-acc} shows the detection accuracy on Dataset~3, which remains above 93\% for both STR models. This result implies that even if the STR pipeline leaves an imperceptible fingerprint, the classifier still accurately captures characteristics specific to STR-ed images, rather than merely detecting whether the image has passed through any neural network layers. \par

\begin{figure}[t]
\centering
\includegraphics[width=0.6\columnwidth]{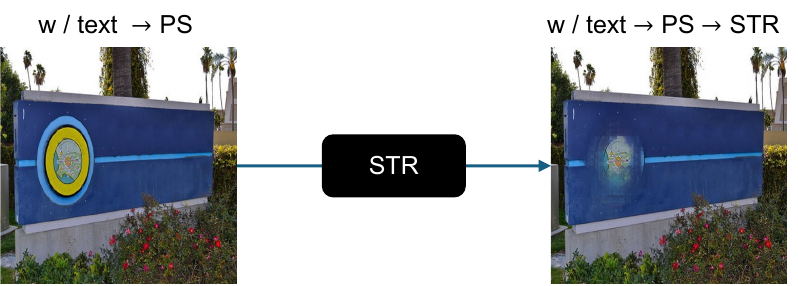}\\[-3mm]
\caption{An example of over-removing a non-text region by ViTEraser. A circular graphic region is removed (maybe by misinterpretation as ``O''). This instance needs to be treated as a negative (i.e., non-STR-ed) instance because the right image is a text-free image.}
\label{fig:nontext-remove}
\end{figure}

On Dataset~3, ViTEraser’s accuracy is about 5\% lower here compared to Dataset~1. ViTEraser sometimes ``over-removes'' non-text regions unnecessarily from originally text-free images. Fig.~\ref{fig:nontext-remove} shows an example of over-removal by STR on a text-free image. Over-removal makes a negative image in Dataset~3 like a positive image, and thus, classification is more difficult; consequently, detection accuracy is reduced compared to the results in the other datasets. 

\subsubsection{Qualitative evaluation}
\begin{figure}[t]
\centering
\includegraphics[width=0.9\columnwidth]{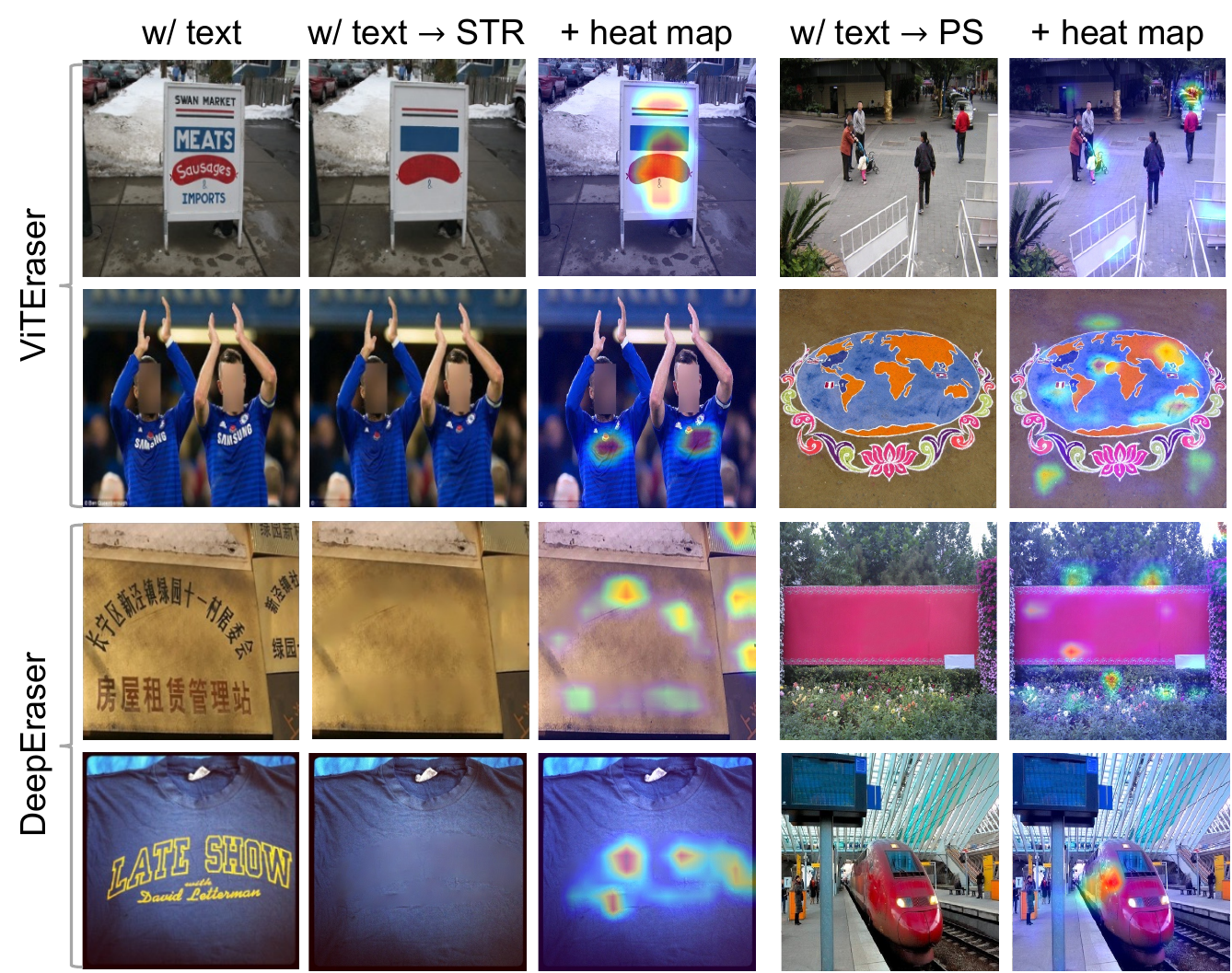}\\[-3mm]
    \caption{Results of ISTR Level-1 (STR presence detection) on Dataset~1. }
\label{fig:level1-results}
\vspace{-5mm}
\end{figure}

Fig.~\ref{fig:level1-results} presents examples of correctly classified images in Level-1 (STR presence detection) for Dataset~1. The three columns on the left show positive instances. In ``w/ text,'' the original image contains visible text; ``w/ text $\to$ STR'' is the same image after STR processing. The ResNet50 binary classifier recognized these latter images as positive instances (i.e., true positives). The ``+heatmap'' column shows Grad-CAM~\cite{selvaraju2017grad} results highlighting the regions to which the classifier paid the most attention. Grad-CAM responses are concentrated in areas where text originally existed, such as signage or sponsor names on a soccer player’s uniform, and it shows minimal reaction to regions without text. This indicates that both ViTEraser and DeepEraser leave an invisible ``fingerprint'' that the classifier exploits for discrimination.

On the right are two columns showing negative instances. ``w/ text$\to$ PS'' represents images where the text was manually removed using Photoshop—these serve as negative instances for Dataset~1. The classifier correctly labeled these images as negative (true negatives). The accompanying ``+heatmap'' reveals the classifier’s attention on various objects, such as strollers, flowers, and leaves, but no clear pattern emerges as in the positive cases. Roughly speaking, it seems to respond more strongly to complex textures (i.e., high-frequency regions) than to smooth areas. Previous research on AI-based content detection has shown that convolutional neural networks can emphasize unique high-frequency components left by manipulation~\cite{sinitsa2024deep,wang2020cnn,wang2023dire}. Our classifier may similarly focus on high-frequency artifacts, distinguishing whether they arise naturally or are generated by STR. 
\par

\section{STR Region Detection\ (ISTR Level-2)}
\subsection{STR Region Detection as a Text Detection Task}
This section focuses on ISTR Level-2, which aims to pinpoint the precise regions of an image that were altered by an STR model. In contrast to Level-1 (STR presence detection), which only determines if an image has been processed by STR, Level-2 delves deeper by localizing the exact areas of text removal. In the STR-ed image shown in Fig.~\ref{fig:ISTR-steps}, texts (``TEA SHOP'' and ``TEA \& SCONES'') have been removed; we investigate whether the detection model can predict their regions (as bounding boxes), even though the texts themselves are no longer visible to human eyes.
\par

Detecting the regions of removed texts is essential because it provides important evidence of how the image was tampered with. For instance, if a detected removed-text region appears on a sheet of paper, it may indicate that crucial document content has been overwritten or hidden. Conversely, detecting a removed region on a car license plate could suggest that the license number was removed for privacy. Moreover, analyzing the size and position of these removed-text regions can reveal the extent and potential intent of the removal. For example, smaller targeted erasures might hint at selective editing for concealment, while larger regions might signal an attempt to eliminate entire blocks of text or paragraphs.
\par

To achieve such localization, we adapt general text detection techniques to the domain of removed text. Specifically, we train a detection model using pairs of STR-ed images and their corresponding original text locations. By learning residual artifacts left behind after text removal, the model can identify removed-text regions in new images. Crucially, if the model detects any inconsistencies in texture or color or any other ``fingerprint'' where text once existed, it can infer the probable bounding boxes of the removed text. 
\par

\subsection{Experimental Setup for ISTR Level-2}
\subsubsection{Dataset and STR models}

\begin{figure}[t]
\centering
    \includegraphics[width=0.5\columnwidth]{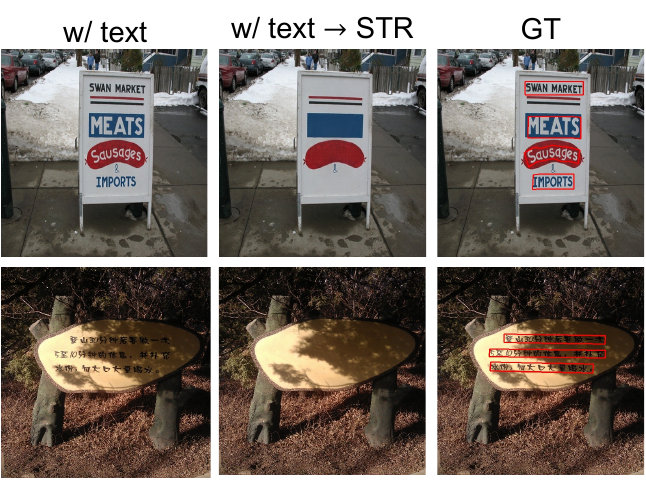}\\[-3mm]
\caption{Images from SCUT-EnsText for the ISTR Level-2 experiments. Ground-truth (GT) indicating text regions are shown as red rectangulars and polygons.}
\label{fig:detection-dataset}
\end{figure}

For our detection dataset, we used SCUT-EnsText~\cite{Erase} along with images processed by an STR model. Fig.~\ref{fig:detection-dataset} shows the several example images. We used the STR-ed images are inputs of the STR region detection model and trained the model with the ground-truth regions, shown in the red polygons in Fig.~\ref{fig:detection-dataset}. Like Level-1, we use DeepEraser~\cite{feng2024deeperaser} and ViTEraser~\cite{peng2024viteraser} as the STR models. 
All 2,749 training images in SCUT-EnsText were fed to DeepEraser or ViTEraser to have 2,749 training instances for the detection model. The 813 images from the dataset were used for testing the performance of the detection model. 

\par

\subsubsection{Evaluation metric}
As a quantitative evaluation metric, we use Intersection over Union (IoU). In this task, each image may include multiple ground-truth (GT) regions (i.e., where text originally existed) and multiple detected regions. Therefore, we compute the IoU by considering all GT regions and all predicted regions within each image.
We use IoU because it can properly evaluate cases where multiple GT regions are detected as a single merged STR region.

\subsubsection{Detection model}

For localizing text-removed regions, we adopt TextBPN-Plus-Plus~\cite{zhang2023arbitrary}, which is a SOTA text detection model and is known for its high accuracy. 
We train TextBPN-Plus-Plus by feeding the STR-processed images (``w/ text$\to$ STR'' in Fig.~\ref{fig:detection-dataset}) as inputs and having the model predict the original text coordinates (``GT''), which now represent text-removed regions. During training for 1,000 epochs, we set the learning rate to 0.001, the batch size to 24, and the optimizer to Adam. We randomly split the 2,749 training images from SCUT-EnsText into 2,199 for training and 550 for validation. The detection model is then trained on those 2,199 images. To select the final detection model for evaluation, we choose the checkpoint that achieves the highest average IoU between the ground-truth removed-text regions and the predicted regions on the 550-image validation set.

\begin{table}[t]
 \caption{STR region detection performance on the test images by IoU $\uparrow$. TextBPN-Plus-Plus~\cite{zhang2023arbitrary} was used as the text detection model.\label{table:STR_detection_eval}}
 \centering
  \begin{tabular}{l|c}
   \hline
    ViTEraser~\cite{peng2024viteraser} & 0.676 \\
    \hline
    DeepEraser~\cite{feng2024deeperaser} & 0.707 \\
   \hline
  \end{tabular}
  \vspace{-3mm}
\end{table}

\subsection{Experimental Results of ISTR Level-2}
\subsubsection{Quantitative evaluation}
Table~\ref{table:STR_detection_eval} shows the average IoU computed on the evaluation set. Both ViTEraser and DeepEraser achieve IoU values around 0.7. Note that many object detection tasks consider IoU~$\ge$~0.5 as a correct detection, and when the target object is as small as text, even minor misalignments can substantially lower IoU. Consequently, obtaining an IoU near 0.7 indicates that the removed-text regions are detected with high precision, suggesting that the ISTR Level-2 task can indeed be accomplished with excellent accuracy.
This result coincides with our Grad-CAM observations in Fig.~\ref{fig:level1-results}, which highlight text regions where STR likely left a ``fingerprint.'' Meanwhile, detection accuracy is slightly lower for ViTEraser. We infer that ViTEraser erases text more aggressively, thereby leaving fewer detectable artifacts and making it harder to localize removed-text regions.

\begin{figure}[t]
    \centering
    \includegraphics[width=\columnwidth]{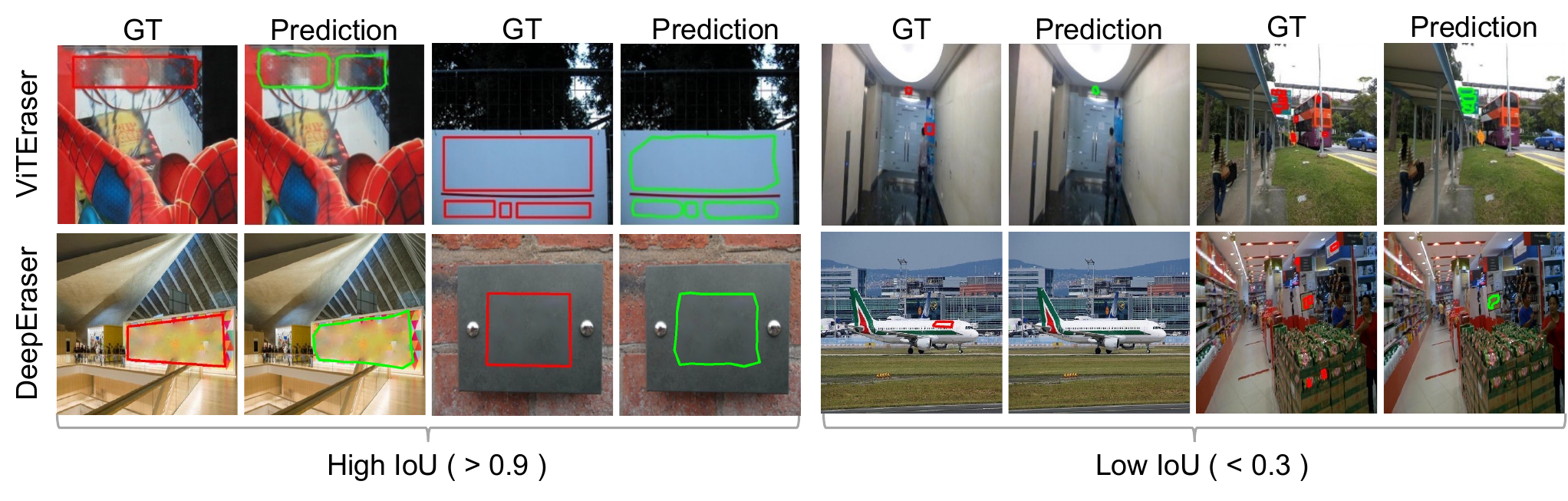}\\[-4mm]
    \caption{Results of ISTR Level-2 (STR region detection) from the test set. }
    \label{fig:level2-results}
    \vspace{-3mm}
\end{figure}

\subsubsection{Qualitative evaluation}
Fig.~\ref{fig:level2-results} shows examples of detection results from our model on STR-ed images. ``GT'' denotes the ground-truth region where text originally existed (shown in red), while ``Prediction'' marks the region identified as STR-ed (shown in green). In the high-IoU examples, the detector accurately locates the previously text-bearing areas ---  including their shapes --- for images STR-ed by ViTEraser or DeepEraser. Notably, this holds true not only when erasure traces are visually apparent, as in the two examples on the leftmost column, but also in cases where text on a signboard was removed so seamlessly that human eyes cannot easily detect it.
\par
The low-IoU examples involve small ground-truth text regions, which tend to give lower IoU. In general, IoU of tiny targets degrades even with a slight displacement. Moreover, our model often misses tiny targets. For instance, in the airplane image, the model does not identify the small removed region on the plane’s body. We infer that when the removed region is extremely small, there may not be enough statistical information to confidently recognize the fingerprint left within that area, ultimately leading to missed detections.

\section{STR Text Recovery\ (ISTR Level-3)} 
\subsection{STR Text Recovery as a Text Recognition Task}

Compared to Level-1 and Level-2, which we showed could be successfully achieved, Level-3 and beyond are anticipated to be far more challenging. In fact, the following preliminary experiments indicate that even Level-3-1 (i.e., recovering the original text within already-known removed-text regions, as shown in Fig.~\ref{fig:ISTR-steps}) proves difficult. ISTR Level-3-1 can be viewed as a text recognition problem on the removed region. Specifically, it is an attempt to ``recognize invisible text by relying on fingerprints.'' By training a text recognizer on images where text has been removed, with the original text as the ground truth, we can empirically verify the feasibility of Level-3-1.

\subsection{Experimental Setup for ISTR Level-3-1}
\subsubsection{Dataset and STR models}
We use SCUT-EnsText along with its STR-ed images. We first need to prepare a pair of a removed-text region and the text originally printed in the region to train a recognition model. 
SCUT-EnsText has ground-truth for text regions (as polygons), but not for text content. Therefore, for images with text before STR (``w/ text'' in Fig.~\ref{fig:detection-dataset}), we apply OCR to text regions where characters exist (``GT'' in Fig.~\ref{fig:detection-dataset}) to prepare our own ``pseudo-ground-truth'' texts. As OCR, we use GOT-OCR2.0~\cite{wei2024generalocrtheoryocr20}, which is a pretrained model known for high-accuracy scene text OCR in English and Chinese. If the recognition result of the text region image shows non-Latin characters and/or uncommon symbols, we excluded the region image from our dataset for Level-3-1. We then pair each STR-ed text-region image (by cropping the STR-ed image) and the pseudo-ground-truth. Following these steps, we generate 15,058 and 4,270 STR-ed text-region images from the SCUT-EnsText training and test sets, respectively, and use them for training and testing the text recognition model for ISTR Level-3-1.

\begin{table}[t]
 
 \caption{Accuracy of STR text recovery (IST Level-3-1). ``Best'' and ``Last'' mean epochs with the highest validation accuracy and end of training. Text-Acc.[\%] and Char-Acc.[\%] are detection accuracies for text-level and character-level.}
 \centering
  \begin{tabular}{l|l|rr|rr}
    \hline
     \multirow{2}{*}{STR model} & \multirow{2}{*}{Split} &  \multicolumn{2}{c|}{Best} & \multicolumn{2}{c}{Last} \\ 
    
    & & Text-Acc. & Char-Acc. & Text-Acc. & Char-Acc. \\
    \hline
    \multirow{3}{*}{ViTEraser~\cite{peng2024viteraser}} & Train & 10.21 & 22.33 & 92.98 & 94.12 \\ 
     & Validation & 2.86 & 8.69 & 3.09 & 8.89 \\ 
     & Test & 2.32 & 8.13 & 2.18 & 7.90 \\
    \hline
    \multirow{3}{*}{DeepEraser~\cite{feng2024deeperaser}} & Train & 89.55 & 91.11 & 90.51 & 91.91 \\
      & Validation & 2.82 & 8.81 & 2.59 & 8.90 \\ 
      & Test & 1.57 & 7.17 & 1.48 & 7.25 \\
   \hline
  \end{tabular}
  \label{tab:level3-1}
\end{table}

\subsubsection{Recognition model}

For the recognition model, we use CLIP4STR~\cite{zhao2024clip4str}, which achieves SOTA performance on a variety of scene Latin-text recognition benchmarks\footnote{CLIP4STR is a SOTA scene text recognizer but cannot recognize Chinese texts by its default setting. We, therefore, did not use it for preparing the pseudo-ground-truth. In contrast, GOT-OCR2.0 can recognize Chinese texts and is suitable for excluding them, although it is not specialized for scene text recognition. Note that we use the CLIP4STR-Large model among several versions of CLIP4STR because of its recognition performance.}. We train it to predict the pseudo-ground-truth text using the STR-ed text-region images. During 200 epochs of training, we set the learning rate to its default value $8.4 \times 10^{-5}$, the batch size to 128, and employ AdamW as the optimizer. From the 15,058 images in the training set, 12,046 were randomly chosen for actual training and 3,012 for validation. We train the recognition model on those 12,046 images and select the final model checkpoint based on the highest text-level accuracy on the 3,012 images for validation.

\subsection{Experimental Results of ISTR Level-3-1}

\subsubsection{Quantitative evaluation}

Table~\ref{tab:level3-1} shows the detection accuracy of the predicted text for train, validation, and test data at ISTR Level-3-1. For finer analysis, we employed two types of detection accuracy metrics: \textit{Text}-Accuracy and \textit{Char}-Accuracy. The former measures the proportion of instances in which the predicted text exactly matches the pseudo-ground-truth text. In contrast, the latter quantifies the similarity between the pseudo-ground-truth text and predicted text at the character level using normalized edit distance\footnote{Normalized edit distance (NED)~\cite{yujian2007normalized} is a similarity measure ranging from 0 to 1, which quantifies the difference between two texts at the character level. A higher NED value indicates greater dissimilarity between the texts. Therefore, we used $(1 - \mathrm{NED}) \times 100\ [\%]$ as Char-Accuracy represents the percentage of character-level similarity.}. Additionally, in Table~\ref{tab:level3-1}, \textit{Best} represents the results obtained using the model weights from the epoch with the highest {\em validation} accuracy, while \textit{Last} represents results from the final epoch of training (i.e., the 200th epoch). 
\par

The ``best'' epoch results in Table~\ref{tab:level3-1} show that regardless of whether the images were STR-ed by DeepEraser or ViTEraser, very low Text- and Char-Accuracies were achieved on the test data. For ViTEraser, both Text- and Char-Accuracies remained low even on the training data. However, from the ``last'' epoch results, we observe that while both DeepEraser and ViTEraser improved their training data accuracy, their validation and test data accuracy remained low, indicating overfitting. These results suggest that even the recent OCR model, CLIP4STR, cannot recognize invisible STR-ed texts. In other words, the recent STR model leaves a fingerprint that indicates whether or not a text has been STR-ed (as we saw in the results of Levels-1 and 2) but does not leave enough of a fingerprint to be able to recover the STR-ed text content.

\begin{figure}[t]
\centering
\includegraphics[width=\columnwidth]{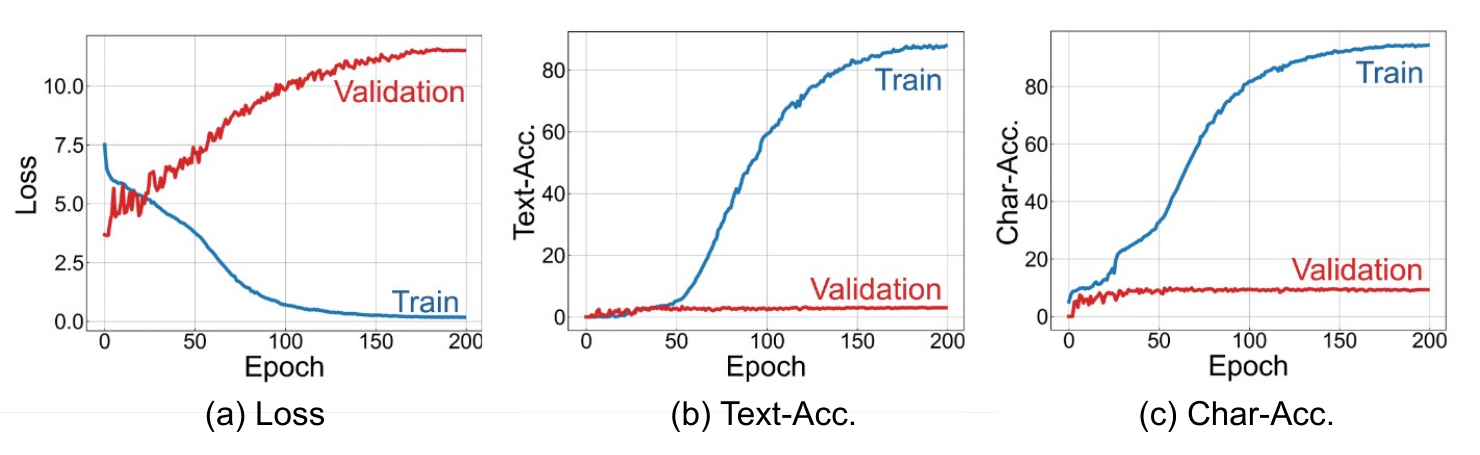}\\[-3mm]
    \caption{Learning curves of ISTR Level-3-1 for training and validation data.}
\label{fig:level3-1-results}
\vspace{-5mm}
\end{figure}

\subsubsection{Analysis of learning curves}

Fig.~\ref{fig:level3-1-results} shows learning curves for the training and validation data on STR-ed images with ViTEraser. From (a), it is evident that while the training loss consistently decreases, the validation loss continues to deteriorate from the early stages of training. Furthermore, as shown in (b) and (c), both Text- and Char-Accuracies in the training data increase and converge to high values toward the later stages of training. In contrast, two accuracies on the validation data remain persistently low throughout the entire training process. 
This finding further reinforces the conclusion that learning for ISTR Level-3-1 remains extremely challenging. The rapid onset of overfitting implies that the model fails to generalize beyond the training data, likely memorizing individual samples rather than capturing meaningful patterns necessary for text reconstruction. Addressing this issue may require alternative training strategies, such as stronger regularization techniques, data augmentation, or novel architectures better suited for the task.

\section{Conclusion} 
In this paper, we investigated the potential of \emph{Inverse} STR (ISTR), focusing on two practical levels: STR presence detection (Level-1) and STR region detection (Level-2). For Level-1, our experimental results demonstrated near-perfect classification accuracy, indicating that even state-of-the-art STR models leave subtle, though invisible, artifacts (called fingerprints). This finding suggests that images processed by STR can be reliably distinguished from natural or manually edited images, thereby supporting the early detection of possible misuse.
In Level-2 experiments, we achieved an average IoU of around 0.7 for localizing removed-text regions, a fairly high value for such a fine-grained task. We anticipated that SOTA STR models would leave fewer artifacts behind due to their visually perfect text-removal performance. Nevertheless, our detection model still captured sizable portions of the removed-text areas, confirming that STR processing does not eliminate all tampering signals.
\par
In contrast, we currently have a pessimistic experimental conclusion about ISTR Level-3 (STR text recovery). Even when we give the removed text region and use a SOTA text recognizer, it was almost impossible to ``recognize'' the invisible removed text in the region. The current result of Level-3 indicates that ISTR Level-4 (STR text restoration) is also intractable. However, it will be worth applying standard image restoration techniques for the removed-text regions in future work. Even though the restored images do not show legible text (or they just show fragments of the original text strokes), they might still be useful as a hint of the original text's appearance.
\par
Another direction for future work is to further improve STR itself through ISTR. STR and ISTR can be thought of as the generator and detector in a GAN-like scenario. Currently, images produced by the generator (i.e., STR) are being detected at Levels 1 and 2 of ISTR. If we feed this detection result back into STR for additional training, STR may learn to remove text in a way that becomes more difficult to detect. Ethically, creating such highly improved STR carries the risk of enabling more sophisticated document tampering. However, as in all security research, improving the attacker’s capabilities can also strengthen the defender’s responses. In other words, improving STR in this manner would simultaneously encourage the development of even more advanced ISTR.

\begin{credits}
\subsubsection{\ackname} 
This work was supported by JSPS KAKENHI JP22H00540, JP25H01149 and JP24KJ1805, and JST CRONOS-JPMJCS24K4.
\end{credits}

\bibliography{refer}
\bibliographystyle{splncs04.bst} 

\end{document}